\documentclass[letterpaper, 10 pt, conference]{ieeeconf}  

\IEEEoverridecommandlockouts                              

\usepackage{graphics} 
\usepackage{mathptmx} 
\usepackage{times} 
\usepackage{amsmath} 
\usepackage{amssymb}  

\usepackage{cite}
\usepackage{graphicx}

\usepackage{hyperref}

\usepackage{soul}

\usepackage[table]{xcolor}

\newcommand{\rev}[1]{{#1}}

\title{\LARGE \bf
Regularizing Action Policies for Smooth Control \\ with Reinforcement Learning
}

\author{Siddharth Mysore$^*$, Bassel Mabsout$^*$, Renato Mancuso and Kate Saenko$^{**}$
\thanks{* Authors contributed equally}
\thanks{All authors are affiliated with the Department of Computer Science, Boston University, Boston, MA 02215. {\tt\scriptsize [sidmys, bmabsout, rmancuso, saenko]@bu.edu}}%
\thanks{** Co-affiliated with MIT-IBM Watson AI Lab}%
}

\begin{document}
\bstctlcite{IEEEexample:BSTcontrol}

\maketitle
\thispagestyle{empty}
\pagestyle{empty}

\begin{abstract}

    A critical problem with the practical utility of controllers trained with deep Reinforcement Learning (RL) is the notable lack of smoothness in the actions learned by the RL policies.
    This trend often presents itself in the form of control signal oscillation and can result in poor control, high power consumption, and undue system wear.
    We introduce Conditioning for Action Policy Smoothness (CAPS), an effective yet intuitive regularization on action policies, which offers consistent improvement in the smoothness of the learned state-to-action mappings of neural network controllers, reflected in the elimination of high-frequency components in the control signal.
    Tested on a real system, improvements in controller smoothness on a quadrotor drone resulted in an almost 80\% reduction in power consumption while consistently training flight-worthy controllers.
    Project website: \url{http://ai.bu.edu/caps}
\end{abstract}

\section{Introduction}
    
\begin{figure}[t!]
        \centering
        \includegraphics[width=0.98\columnwidth]{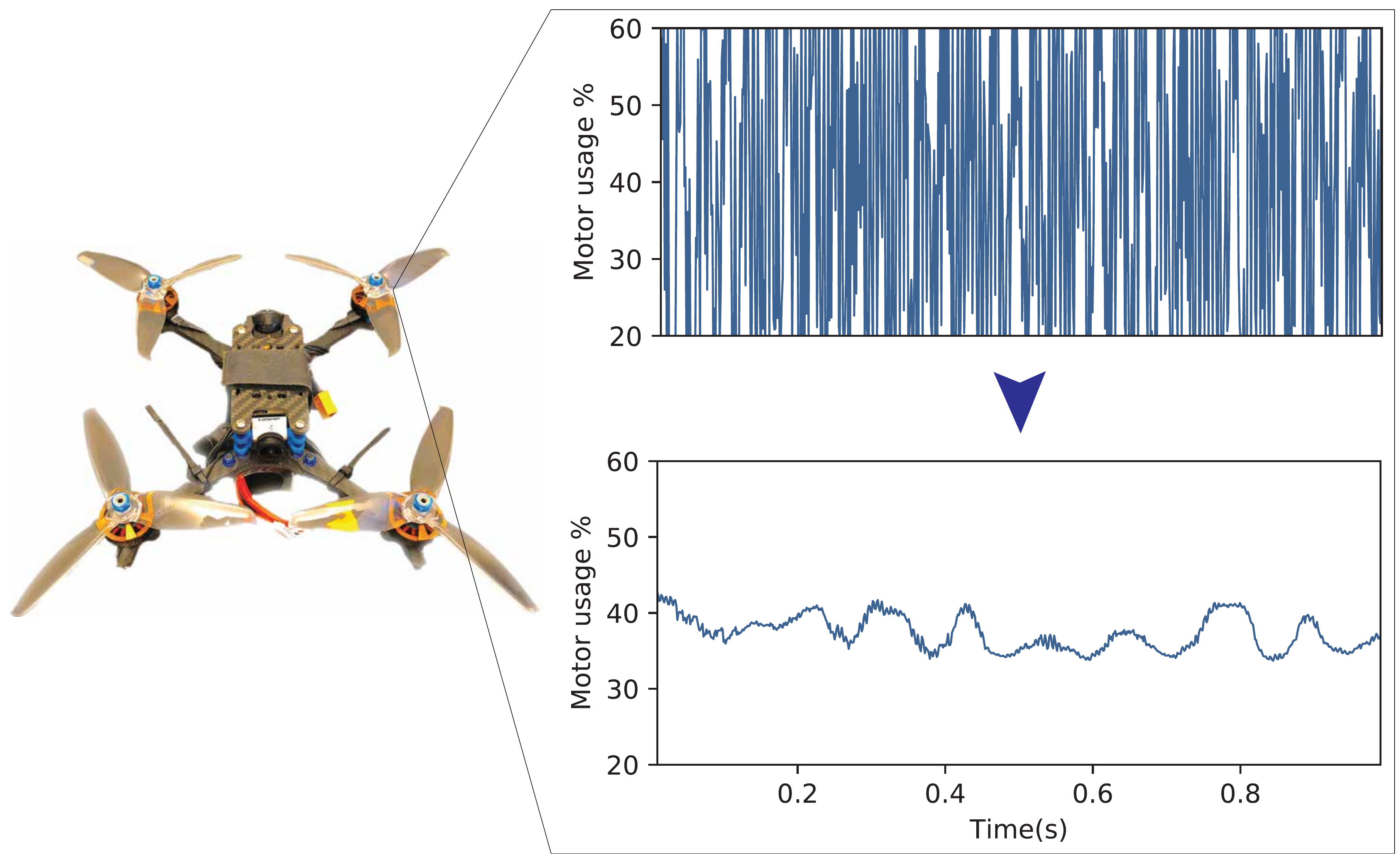}
        \vspace{-.5\baselineskip}
        \caption{We introduce Conditioning for Action Policy Smoothness (CAPS), a regularization tool that encourages RL agents to learn fundamentally smoother control policies. A state-of-the-art neural network controller (top), trained in simulation with highly tuned rewards, still produces a noisy motor control signal when deployed on our physical drone platform. Training with CAPS (bottom) using the same pipeline and simpler rewards yields significantly smoother control in contrast.}
        \label{IntroFig}
        \vspace{-1.65\baselineskip}
    \end{figure}

    A key advantage of deep Reinforcement Learning (RL) is that deep RL is capable of training control policies for complex dynamical systems for which it would be difficult to manually design and tune a controller~\cite{Dextrous, RoboImitationPeng20, mnih2015human, DDPG, benchmarkingRobo}.
    RL-based controllers may however exhibit problematic behavior that is detrimental to the integrity of the controlled system~\cite{benchmarkingRobo, Sim2Real, Sim2multi, benchmarkingRL, NFThesis}. 
    A critical issue observed both in literature and from our tests, but one that has not received much attention,
    is with oscillatory control responses (\rev{example:} Fig.~\ref{IntroFig}). 
    This \rev{may} have undesirable consequences ranging from visible oscillations in the physical system response~\cite{Sim2multi, NFv2}, increased power consumption or over-heating~\cite{benchmarkingRobo, NFThesis} due to high-frequency oscillations, and even hardware failure~\cite{benchmarkingRobo}.
    
    \rev{This problem is particularly pronounced in continuous control, where controller response can vary infinitely within the limits of acceptable outputs, and can be further accentuated by the domain gap in sim-to-real transfer.}
    Yet, training controllers in simulation is attractive because it alleviates concerns of hardware and user safety.
    In this work, we focus on deep RL-based low-level attitude controllers for high-performance quadrotor drones.
    Specifically, we train controllers to follow a pilot's input desired rates of angular velocity on the three axes of rotation of the drone.
    
    The black-box nature of neural network-based (NN) controllers limits mitigation of bad behavior \rev{at} run-time. 
    \rev{Filters can} address problems of oscillatory control in classical control systems, but our experiments showed that filters \rev{behaved inconsistently with NN controllers}, 
    \rev{requiring the problem to be addressed} at the control policy level \rev{during training}.
    RL assumes black-box training environments, so reward signals are used as a stand-in for a differentiable measure of agent performance.
    Attempts to condition the behavior of RL agents, i.e. the trained control policies, largely focus on engineering rewards to induce the desired behavior~\cite{Sim2multi,benchmarkingRobo, NFThesis, Hwangbo2017ControlOA}.
    \rev{However,} relying on reward-engineering for behavior conditioning requires networks to learn through indirect information, which may or may not be easily relatable to the outputs of the network.
    The process is tedious, unintuitive and ultimately provides no guarantees that the \rev{desired} behavior \rev{would be} learned~\cite{Ilyas2020A}.
    For some behavior traits, the need for reward-engineering may be unavoidable; but where it is possible to define desirable behavior traits with respect to the actions taken by a RL agent's action policy, we posit that it is better to directly optimize the action policy, instead of relying on reward engineering.
    Achieving smooth control policies, as we demonstrate, is one such case.
    
    We propose the ``Conditioning for Action Policy Smoothness" (CAPS) regularization constraint that promotes two aspects of action policy smoothness: (i) temporal smoothness, i.e. actions taken should be similar to any previous actions to preserve smooth transitions between controller outputs over time, and (ii) spatial smoothness, i.e. similar states map to similar actions, thus mitigating measurement noise and modeling uncertainties.
    Optimizing policies with CAPS enables consistent training of smooth action policies.
    
    CAPS is intuitive yet highly effective in improving policy smoothness.
    Its relative simplicity also allows CAPS hyperparameters and the agents' response to them to be {more explainable, thus more easily tuned.}
    To facilitate analysis, we also propose a metric with which to evaluate the smoothness of trained policies.
    CAPS provides consistent demonstrable improvement in smoothness on a number of OpenAI continuous control benchmarks.
    Additionally, CAPS regularization consistently trained flight-worthy drone controllers despite using a simple reward structure --- something that previously failed to produce flight-safe agents; it achieved an almost 80\% reduction in power consumption compared to the state of the art, with 96\% improvement in policy smoothness, and significantly reduced training time.
    \rev{While CAPS did sometimes contribute to marginally worse training rewards due to controllers' responses being more conservative, this was never at the cost of learnability or practical utility.}

\section{Background}\label{Background}

    Smooth policy response is not a problem that has received much attention from the broader research community, despite being an important consideration for practical application of RL. 
    As no aspect of the standard RL training formulation inherently accounts for policy smoothness, prior works that highlight issues of control smoothness attempt to encourage more robust behavior by engineering the rewards received during training~\cite{Sim2multi, benchmarkingRobo, NFThesis}.
    These approaches have two drawbacks: (i) they require that the smoothness be captured numerically in the reward signal, which may require access to information about the state of the environment that may not be readily accessible, and
    (ii) RL training environments are often black boxes where the functions mapping policies to rewards, i.e. the \emph{true} value functions, are unknown. Thus, RL algorithms simultaneously learn a \emph{surrogate} value function to evaluate the quality of a policy's behavior.
    
    Research has shown that RL algorithms can be very sensitive to learning hyper-parameters and initialization~\cite{Overfitting, Overfitting2}, as well as input normalization, reward scales, and code optimizations~\cite{Engstrom2020Implementation}.
    This has resulted in serious repeatability issues in RL literature~\cite{Overfitting, Overfitting2, pineau2020improving}.
    Ilyas et al.~\cite{Ilyas2020A} demonstrated that the surrogate optimization functions learned by policy gradient algorithms can present a very different optimization space when compared to the true value function, such that optimizing for the surrogate can even result in worse policy performance as training progresses.
    It is therefore difficult to put trust in nuanced information transfer from environment rewards to action policy optimization.
    We instead optimize policies directly.
    
    Action-smoothness is purely a function of the actions taken by the trained policy and can be conditioned using regularization penalties in policy optimization. 
    Prior works have explored the value of regularization techniques in RL.
    Liu et al.~\cite{liu2019regularization} highlight regularization as an understudied tool in RL applications and demonstrated the utility of regularization in improving policy performance.
    They however regularize the weights of the neural networks for improving performance on reward metrics, but do not address the differences in behavior that results from regularization.
    Conversely, we focus on regularizing the action policy mappings, with specific attention to the characteristics of the learned behavior.
    Work in representation learning by Jonschkowski \& Brock~\cite{repRL, repRL2} shows that using temporal coherence constraints in learning state representations can improve RL in robotics applications.
    We apply a similar idea to action mappings as opposed to state mappings.
    Concurrent with our efforts, Shen et al.~\cite{shen2020deep} developed a technique for improving the robustness of RL agents to noise by, in effect, regularizing policies for spatial smoothness, though the specific mathematical formulation and implementation differs from ours. 
    While the authors are able to demonstrate increased performance on standard reward metrics and an improved robustness to noise, they do not provide metrics on the behavior of the learned policies and whether they are indeed smoother in their mappings. 
    For our problem of minimizing oscillatory responses in learned control, particularly when transferring controllers from simulated training environments, we anticipated that conditioning for spatial smoothness alone would not be enough, which is why we developed our method with both spatial and temporal smoothing in mind.

\section{CAPS : Conditioning for Action Policy Smoothness}\label{Methdo}

    We propose `Conditioning for Action Policy Smoothness' (CAPS), a simple and effective tool for conditioning action policies for smooth behavior through regularization.
    It is typical to condition policy behavior through reward signals which are then approximated by the value function. 
    We instead chose to focus on addressing policy behavior at the source --- the neural network which represents the action policy and condition it for smooth actions.
    
    To condition policies for smooth behavior by directly affecting how smoothly states are mapped to actions, we propose two regularization terms: (i) a temporal smoothness term requiring subsequent actions taken under the policy to be similar to the action taken on the current state, and (ii) a spatial smoothness term requiring similar states to map to similar actions.
    We hypothesize that, while temporal smoothness would likely be sufficient for the desired control smoothness in well-modeled environments, spatial smoothness would be important in improving policy robustness to domain shifts and unmodeled dynamics, thus allowing for more effective transfer from simulation to real control.
    Empirical evidence in Section~\ref{Eval} supports this hypothesis. 
    
    Many commonly used and state-of-the-art deep RL algorithms formulate optimization with two primary components: (i) optimizing the agent policy and (ii) optimizing the surrogate value function.
    CAPS focuses on how the policy is optimized without directly affecting how the value functions are learned.
    Despite the representational limitations of surrogate value functions, practically, they still provide a useful metric to optimize policies on, so we leave improvements to value estimation as future work.
    CAPS subverts issues that may arise from surrogate functions failing to accurately represent the optimization space, relying less on carefully constructed reward signals.
    Furthermore, since the regularization metrics are based entirely on how the policy networks map states to actions, there is also no reliance on additional information from the environment during training.
    
    RL policies, $\pi$, are functions that map states, $s$, to actions $a = \pi(s)$.
    RL attempts to train an optimal policy $\pi^*$ to maximize expected cumulative rewards, $R$, for all trajectories, $\tau_{\pi^*}$, sampled under policy $\pi^*$.
    Deep RL methods typically build upon policy gradient methods~\cite{sutton2000policy} for solving the RL problem.
    Typically, a neural network policy $\pi_\theta$ is parameterized by $\theta$, with a performance measure on the policy, $J(\pi_\theta)$. 
    Parameters $\theta$ are optimized to maximize $J$. 
    In algorithms based on Q-learning~\cite{Qlearning}, e.g., Deep Deterministic Policy Gradient (DDPG)~\cite{DDPG}, Twin Delayed DDPG (TD3)~\cite{TD3} and Soft Actor Critic (SAC)~\cite{SAC}, the policy optimization function, $J_{\pi_\theta}$, is proportional to the Q-value:
    \begin{align*}
        J_{\pi_\theta} &\propto Q_{\pi_\theta}(s,\pi_\theta(s)) = \mathbf{E}_{\tau_{\pi_\theta}}[R(\tau_{\pi_\theta}) | s_0 = s, a_0 = a]\\ &\propto \mathbf{E}_{s'} [r(s,a) + \gamma Q_{\pi_\theta} (s',\pi_\theta(s'))]
    \end{align*}
    where $s'$ is the next state reached, $r(s,a)$ defines the per-step reward, and $\gamma$ is the discount factor.
    Algorithms based on the more standard policy gradient optimization, such as Trust Region Policy Optimization (TRPO)~\cite{TRPO} and Proximal Policy Optimization (PPO)~\cite{PPO} use the advantage function, $A_{\pi_\theta}$, and a measure of policy divergence:
    \begin{align*}
        J_{\pi_\theta} &\propto \log(\pi_\theta(a | s)) A_{\pi_\theta}(s,a)\\ &\propto \log(\pi_\theta(a | s)) \left(Q_{\pi_\theta}(s,a) - V_{\pi_\theta}(s)\right)\\
        \text{where } V_{\pi} (s) &= \mathbf{E}_{\tau_{\pi}}[R(\tau_{\pi}) | s_0 = s] = \mathbf{E}_{s'} [r(s,\pi(s)) + \gamma V_{\pi} (s')]
    \end{align*}
    
    We construct the CAPS optimization criteria, $J_\theta^{CAPS}$, with a temporal smoothness regularization term, $L_T$, and spatial smoothness term, $L_S$, along with regularization weights, $\lambda_T$ and $\lambda_S$, which control the strength of the regularization terms.
    $L_T$ and $L_S$ are computed using their corresponding distance measures, $D_T$ and $D_S$.
    At time, $t$:
    \begin{gather}
        J_{\pi_\theta}^{CAPS} = J_{\pi_\theta} - \lambda_T L_T - \lambda_S L_S \label{J_CAPS}\\
        L_T = D_T\left({\pi_\theta}(s_t), {\pi_\theta}(s_{t+1})\right)  \label{L_T}\\ 
        L_S = D_S\left({\pi_\theta}(s_t), {\pi_\theta}(\bar{s_t})\right) \ \ \ \text{where } \bar{s} \sim \phi(s_t) \label{L_S} 
    \end{gather}
    $L_T$ penalizes policies when actions taken on the next state under the state transition probabilities of the system are significantly dissimilar from actions taken on the current state.
    $L_S$ mitigates noise in system dynamics by ensuring that policies take similar actions on similar states, $\bar{s}$, which are drawn from a distribution $\phi$ around $s$.
    \rev{
    In effect, these minimize the temporal and spatial Lipschitz constants of the policy functions around regions of interest.
    While computing and optimizing exact Lipschitz constants for neural networks has been shown to be NP-hard~\cite{scaman2018lipschitz}, regularization techniques allow for an approximation with demonstrable utility in increasing the generalizability and robustness of learned mappings~\cite{miyato2018spectral, cisse2017parseval}, though prior exploration in their applicability to RL-based control has been limited.
    }
    
    Distance measures, $D_T$ and $D_S$, and sampling function $\phi$ should be selected appropriately for the data-space used.
    In practice, we use Euclidean distances, i.e. $D(a_1, a_2) = ||a_1 - a_2||_2$, \rev{which made sense as a similarity measure for actuator outputs for the considered tasks}, and assume a normal distribution, $\phi(s) = N(s,\sigma)$, with standard deviation, $\sigma$, based on expected measurement noise and/or tolerance.

    \begin{table*}[t]
      \caption{Comparing rewards and smoothness scores on OpenAI Gym benchmarks}
      \label{bench-results-table}
      \centering
      \setlength\tabcolsep{3.3pt}
      \begin{tabular}{c|c|c|| c|c|| c|c|| c|c}
        \hline
          & \multicolumn{8}{c}{Environment} \\ \cline{2-9} 
        Algorithm   &  \multicolumn{2}{c||}{Pendulum-v0}     &      \multicolumn{2}{c||}{LunarLanderContinuous-v2}   & \multicolumn{2}{c||}{Reacher-v2} & \multicolumn{2}{c}{Ant-v2}  \\ \cline{2-9}
          & Reward $\uparrow$ & Sm $\cdot 10^{3}$ $\downarrow$ & Reward $\uparrow$ & Sm $\cdot 10^{3}$ $\downarrow$ & Reward $\uparrow$ & Sm $\cdot 10^{3}$ $\downarrow$ & Reward $\uparrow$ & Sm $\cdot 10^{3}$ $\downarrow$ \\ \hline
        DDPG~\cite{DDPG} & $ -145.56 \pm 10.64$ & $47.6 \pm 10.64$   & $ 217.04 \pm 51.61$ & $34.9 \pm 1.36$     & $ -4.26 \pm 0.25 $ & $4.56 \pm 0.45$     & $225.23 \pm 362.88$      & $2.73 \pm 0.65$ \\
        DDPG + CAPS      & \cellcolor{red!30} $-188.16 \pm 22.53$     & \cellcolor{green!85} $7.09 \pm 1.65$ & \cellcolor{red!17} $181.98 \pm 87.18$     & \cellcolor{green!52} $ 16.7 \pm 2.92$ & \cellcolor{red!19} $-5.03 \pm 1.89$      & \cellcolor{green!19} $ 3.69 \pm 1.13$ & \cellcolor{green!12} $ 253.30 \pm 187.93$  & \cellcolor{green!52} $ 1.31 \pm 0.71$ \\ \hline \hline
        SAC~\cite{SAC}   & $ -139.86 \pm 8.29$  & $9.32 \pm 1.13$    & $277.62 \pm 11.02$     & $8.14 \pm 0.81$     & $-5.96 \pm 0.47$   & $5.99 \pm 0.91$     & $3366.07 \pm 1522.45$    & $6.53 \pm 2.26$ \\
        SAC + CAPS       & \cellcolor{red!19} $-165.79 \pm 9.22$      & \cellcolor{green!47} $4.93 \pm 1.15$ & \cellcolor{green!1} $281.94 \pm 3.65$   & \cellcolor{green!6} $7.62 \pm 0.71$  & \cellcolor{red!5} $-6.25 \pm 3.71$      & \cellcolor{green!16} $ 5.00 \pm 0.71$ & \cellcolor{green!25} $4209.08 \pm 1367.18$ & \cellcolor{green!6} $6.11 \pm 2.93$ \\ \hline \hline
        TD3~\cite{TD3}   & $-152.71 \pm 9.47$   & $43.9 \pm 30.94$   & $271.06 \pm 17.39$  & $37.9 \pm 12.30$    & $-6.52 \pm 1.12$      & $5.70 \pm 0.98$     & $3087.86 \pm 888.75 $    & $9.09 \pm 1.90$ \\
        TD3 + CAPS       & \cellcolor{red!14} $-172.82 \pm 13.47$     & \cellcolor{green!86} $5.92 \pm 1.52$ & \cellcolor{red!1} $270.32 \pm 25.73$     & \cellcolor{green!55} $16.7 \pm 3.26$  & \cellcolor{green!2} $ -6.34 \pm 0.66$  & \cellcolor{green!18} $ 4.63 \pm 0.72$ & \cellcolor{green!25} $3871.68 \pm 1121.36$ & \cellcolor{green!13} $7.89 \pm 2.92$ \\ \hline \hline
        PPO~\cite{PPO}   & $-668.60 \pm 551.85 $   & $9.29 \pm 5.51$    & $ 169.08 \pm 56.59$ & $11.4 \pm 1.36$     & $ -4.37 \pm 1.74$  & $4.49 \pm 0.43$     & $3734.58 \pm 988.29$     & $6.09 \pm 1.19$ \\
        PPO + CAPS       & \cellcolor{green!11} $-590.35 \pm 295.86$ & \cellcolor{green!12} $8.09 \pm 2.13$ & \cellcolor{red!17} $140.71 \pm 23.03 $    & \cellcolor{green!12} $10.0 \pm 2.92$  & \cellcolor{red!8} $-4.69 \pm 2.05$      & \cellcolor{green!24} $ 3.38 \pm 0.36$ & \cellcolor{green!13} $4256.93 \pm 570.88$  & \cellcolor{green!73} $1.60 \pm 0.26$ \\ \hline
        \end{tabular}
        \vspace{-1.8\baselineskip}
    \end{table*}
    
\section{Evaluation}\label{Eval}

    We analyze the effectiveness of CAPS in three phases. 
    First, we verify that it works as expected on a simple toy problem where the theoretically optimal control is known. 
    Following this, we explore its broader applicability on a set of standard benchmark tasks provided by OpenAI Gym~\cite{GYM}. 
    Finally, we apply CAPS to the control of a quadrotor drone (pictured in Fig.~\ref{IntroFig}) to empirically demonstrate the practical utility of our proposed tools and compare its performance against a prior attempt at achieving smooth control through reward engineering~\cite{NFThesis}.
    Implementation details, code and full test results are provided at \url{http://ai.bu.edu/caps}.
    
    \begin{figure}[t]
    	\centering
        \includegraphics[width=.95\columnwidth]{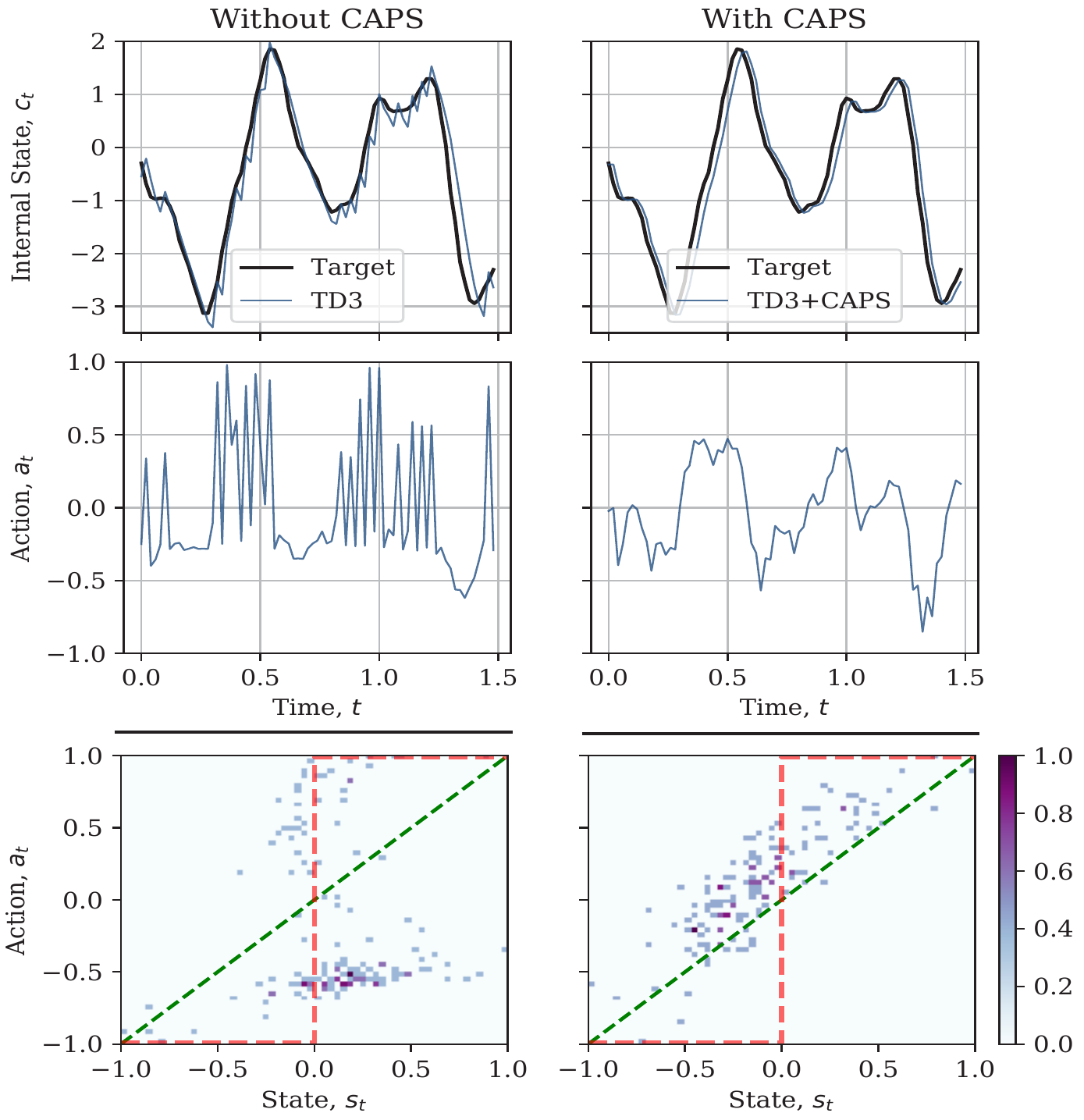}
        \vspace{-.8\baselineskip}
     	\caption{Comparison of the state response and normalized state-action histograms on the Toy problem for a TD3 policy (left) vs a CAPS-regularized TD3 policy of the same architecture (right). Note that actions of the regularized agent are much closer to the ideal linear mapping (green dotted line) which would optimally solve this toy problem while the actions of the vanilla agent are closer to a binary policy (red dotted line). This corresponds with the high-frequency oscillation observed in the control. For illustrative clarity, we only show TD3 though the other RL algorithms tested had similar responses.}
        \label{ToyFig}
        \vspace{-1.5\baselineskip}
    \end{figure}
    
    \paragraph*{\textbf{Toy Problem}}
        To illustrate that issues related to smooth control are not limited to applications of RL with complex dynamics, we constructed a simple 1-dimensional goal-tracking environment with no complex dynamics or noise. 
        The agents observe the disparity, \rev{$s_t = g_t – c_t$}, between the current state, \rev{$c_t$}, and the desired goal state, $g_t$. 
        The actions the agent takes directly affect the system response such that \rev{$s_{t+1} = c_t + a_t$}. 
        It is clear to see that the ideal action would be \rev{$a_t^* = s_t$}. 
        We observed that RL agents tended to learn highly aggressive control policies, akin to a binary step response, which strove to achieve the goal states as quickly as possible, as shown in Fig.~\ref{ToyFig}. 
        This results in oscillations in the action space of the agent as it tries to maintain good tracking. 
        It also seems that oscillatory behavior learned early in training acts as a local minimum that the agents fail to break out of. 
        When employing CAPS however, the learned behavior is much closer to the ideal, naturally yielding smoother control.

    \paragraph*{\textbf{Measuring Smoothness}}
        Smoothness is easily observed visually for a 1-D toy problem, but is more difficult to generally quantify for multi-dimensional action spaces.
        When analyzing the quality of control signals, prior work~\cite{Sim2multi} and engineering tools for analyzing control signals~\cite{betaflight-homepage} have measured outlier peaks of high-amplitude and high-frequency in the control signal spectrum. 
        We observed however that neural network control signals did not always have distinct peaks in the frequency spectrum but would sometimes present with non-trivial amplitudes of frequency components across the spectra, thus making the notion of a `peak' frequency a poor measure for evaluating smoothness. 
        
    
        In order to objectively measure and compare smoothness, we define a smoothness measure, $Sm$, based on the Fast Fourier Transform (FFT) frequency spectrum:
        \begin{equation}
            Sm = \frac{2}{n f_{s}} \sum_{i=1}^n M_i f_i
            \label{smoothness_eq}
        \end{equation}
        where $M_i$ is the amplitude of $i^{\text{th}}$ frequency component, $f_i$ for $i \in [1, n]$ and $f_s$ is the sampling frequency (note here that the maximum resolvable frequency component of the control signal spectrum is the Nyquist frequency, i.e. $\frac{1}{2}f_s$).
        By jointly considering the frequencies and amplitude of the control signal components, this metric provides the mean weighted normalized frequency. 
        For a given control problem, higher numbers on this scale indicate the presence of larger high-frequency signal components and is typically predictive of more expensive actuation, while lower numbers indicate smoother responses.
        Much like comparing rewards however, comparing smoothness numbers between different problems is not necessarily meaningful.
        We demonstrate that CAPS allows for policies to achieve comparable, if not better performance than their unregularized counterparts, while learning smoother action policies on all problems tested.

    \paragraph*{\textbf{Gym Benchmarks}}\label{bench_eval}
        We evaluated \rev{CAPS} on 4 simulated benchmark tasks from OpenAI's gym~\cite{GYM} with 4 commonly used RL algorithms. 
        Results are shown in Table~\ref{bench-results-table}, representing the mean rewards and smoothness scores achieved by {11} independently trained agents in each case.
        Hyperparameters for training were based on those provided in the stable-baselines zoo~\cite{stable-baselines}.
        An ablation study \rev{(available on our website)} helped determine good regularization parameters for each algorithm and environment. 
        Codes were based on OpenAI's Spinning Up~\cite{SpinningUp2018}, with the exception of PPO, which uses the OpenAI Baselines~\cite{baselines} implementation.
        
        CAPS agents are smoother \rev{than their vanilla counterparts} on all tested tasks.
        We observe a nominal performance hit on the pendulum and lunar-lander tasks on numerical rewards, owing to agents being slower at achieving goal states as they strive for smoother behavior, which is in turn reflected in the significantly improved smoothness scores.
        With the Reacher and Ant environments, gains in smoothness were limited inherently by the system dynamics. 
        Interestingly, soft-policies such as PPO and SAC appear to learn relatively smoother policies on their own. 
        We hypothesize that the cause is correlated to stochasticity in the policies allowing for improved exploration of the state and action spaces. 
        
    
    \paragraph*{\textbf{Sim-to-Real Policy Transfer}}\label{drone_eval}
        
        Where the benchmark tasks demonstrate the general utility of CAPS, testing on our drone hardware demonstrates its practical utility on real problems.
        The framework for constructing the drone, and for training and deploying trained networks to the drone builds from the Neuroflight architecture developed by Koch et al.~\cite{NFThesis, NFori, NFv2}.
        The main problem with Neuroflight is that, despite achieving decent tracking of the control-input in simulation, the trained agent presented with significant high-frequency control signal oscillations, which would cause the drone's motors to overheat or even make the drone un-flyable.
        Moreover, the increased power draw also resulted in limited flight-time.
        We suspect that policies are overfit to the complicated rewards on the simulated dynamics, and are compromised by the domain shift when transferring policies to real hardware.
        Furthermore, we noted issues with reproducibility. 
        Only a handful of agents were flight-worthy, while most agents failed to transfer well to the real drone.
        Consequently, while the Neuroflight framework offers a good basis to develop neural network-based control in resource-constrained practical applications, the limited reproducibility and undesirable network behavior in transfer offer large room for improvement.
        For our experiments, we use the same PPO algorithm, network architecture and deployment pipeline as Koch et al. and compare against their best-trained agent, which we acquired from the authors.
        
        \begin{figure}[t]
          \centering
          \includegraphics[width=\columnwidth]{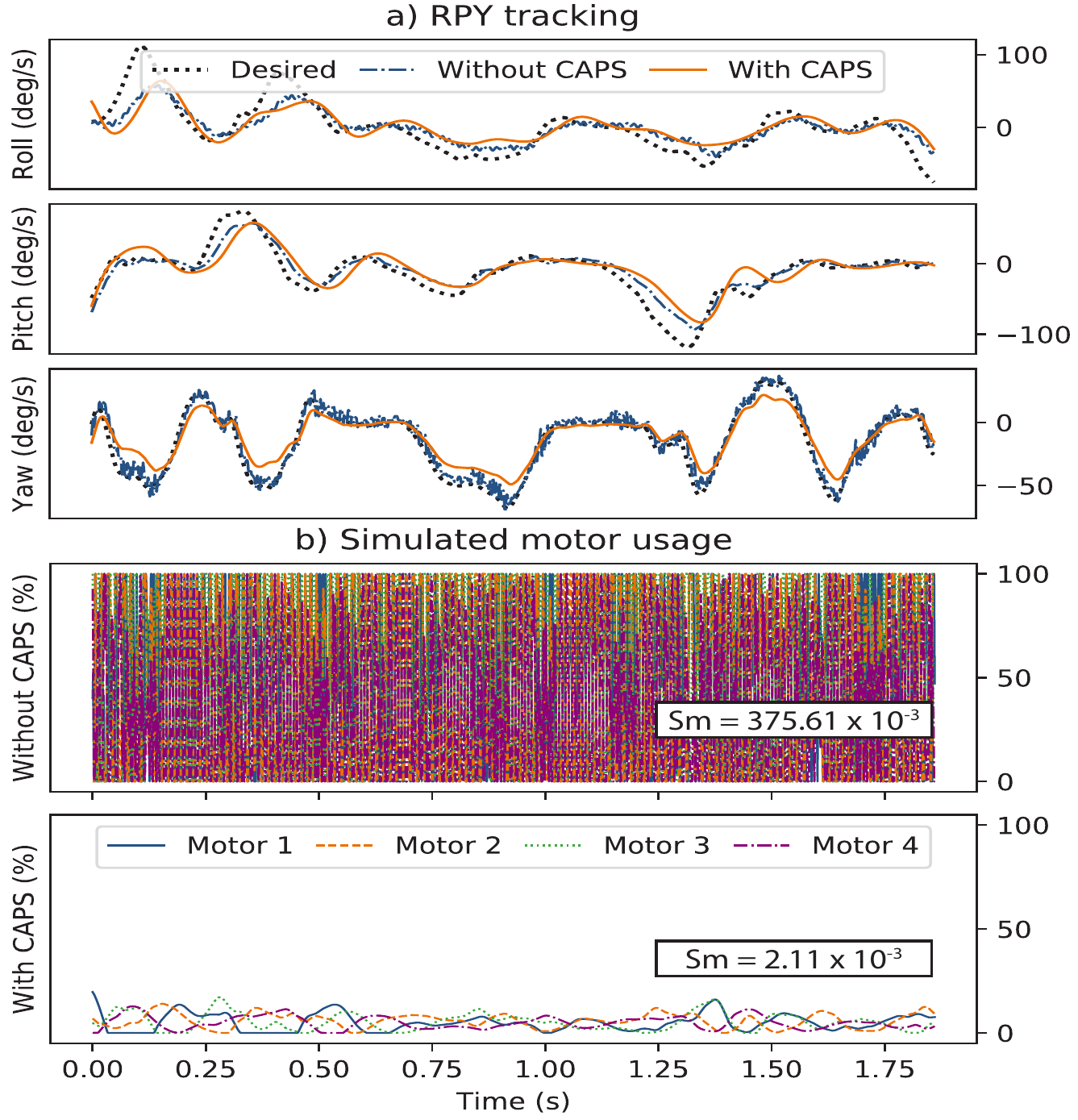}
          \vspace{-1.8\baselineskip}
          \caption{PPO trained for quadrotor control with just a tracking error reward compared against the same algorithm with CAPS optimization. Note the significant reduction in motor signal amplitude and oscillation, despite maintaining similar tracking performance. When trained with CAPS we noted that trained agents were in fact flight-worthy --- i.e. would not pose a significant risk to the pilot or hardware to test. For actual flight, agents had to be conditioned to maintain a minimum output throttle in order to integrate properly with the throttle mixer used by the firmware.}
          \label{Drone-err}
          \vspace{-1.8\baselineskip}
        \end{figure}
        
        \begin{figure*}[t]
          \centering
          \scalebox{1}[0.9]{
          \includegraphics[width=\textwidth]{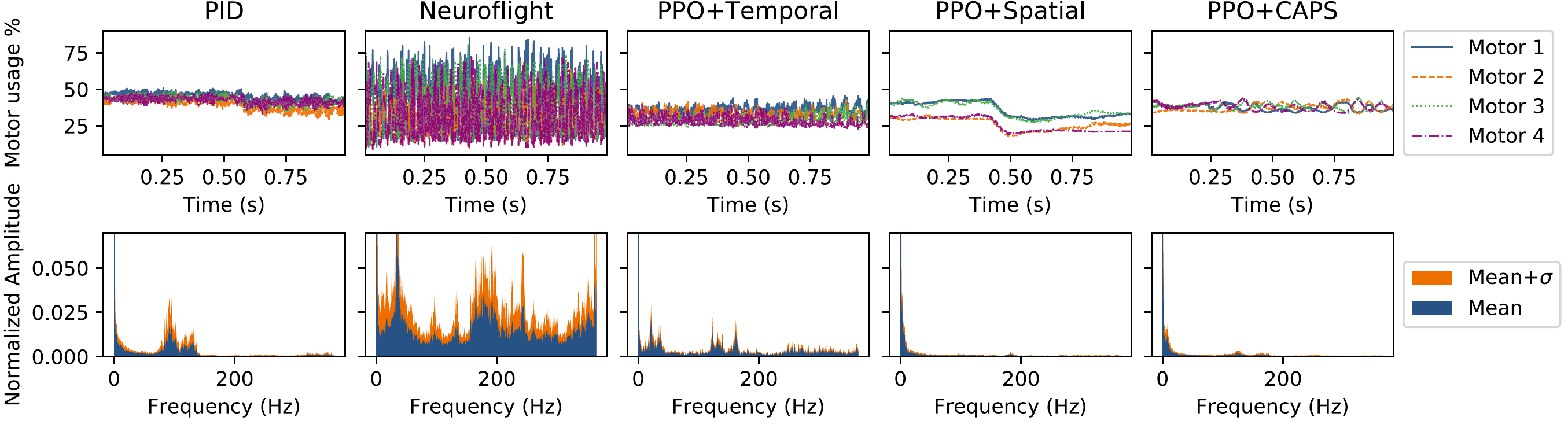}}
          \vspace{-1.8\baselineskip}
          \caption{
            Comparison of motor usage during flight and corresponding FFTs demonstrate the significant improvement in smoothness of control signals with CAPS, with high frequency components being practically eliminated. This contrasts strongly with Neuroflight, which despite being conditioned for smooth tracking, quickly falls victim to the domain gap between the dynamics of the simulated training environment and real flight. CAPS however demonstrates robustness to this shift, which likely contributes to the fact that all agents trained with CAPS were flight-worthy, despite using the same architecture and general pipeline as Neuroflight, whereas the latter required cherry-picking of viable controllers. We also show how temporal and spatial smoothing do not fully address the problem of smoothness, with temporally smoothed control being relatively noisy in transfer and spatially smoothed control forcing control signals into bands of control. CAPS is able to draw the benefits of both without the same drawbacks.
          }
          \label{Drone-FFT}
        \end{figure*}
        
        \begin{figure*}[t]
          \centering
          \scalebox{0.95}[0.9]{
          \includegraphics[width=\textwidth]{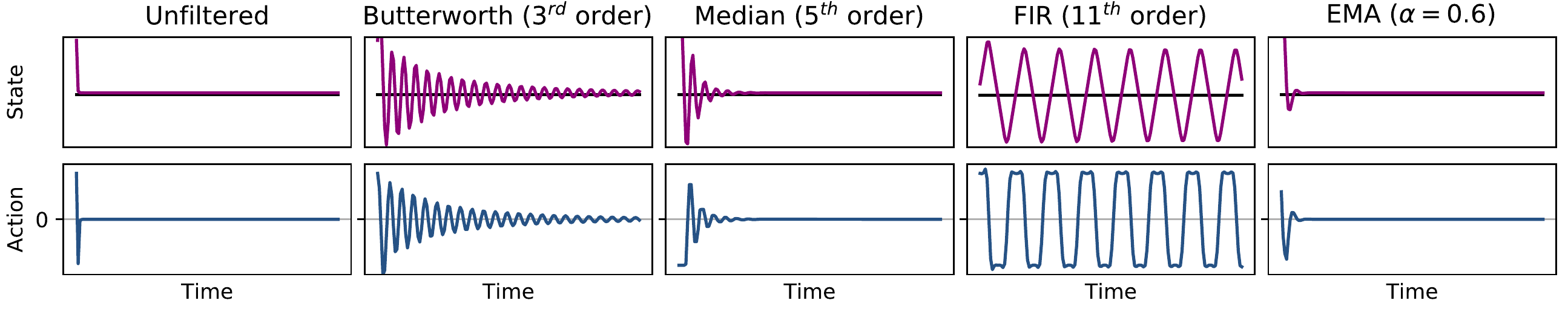}}
          \vspace{-1.\baselineskip}
          \caption{
              We demonstrate the potential negative impact of filtering by testing a number of filters on the action outputs of a well-trained neural network controller tracking a steady-state input on our simple Toy problem.
              The filters were tuned for critical damping on a tuned PID controller.
              Where classical controllers would quickly adjust their outputs in response to the filtering, neural networks may have an over-fit model for system dynamics and may attempt to over-compensate for unfamiliar dynamical response.
              This can result in minor overshoot and oscillation as with the Median and Exponential Moving Average~(EMA) filters, but could also result in a total and catastrophic loss of control as seen with the Finite Impulse Response~(FIR) filter.
              Additional examples of the misbehavior of filtering with neural network-based controllers can be found at our website \url{http://ai.bu.edu/caps}.
          }
          \label{FilterToy}
          \vspace{-1.3\baselineskip}
        \end{figure*}

        Neuroflight uses heavily engineered training rewards to achieve smooth control in simulation.
        By stripping the training rewards down to their basic tracking error components and comparing agents trained with CAPS against those without, we got a clearer sense of how our proposed regularization affects the learned policies in the absence of any additional conditioning.
        As shown in Fig.~\ref{Drone-err}, agents trained with CAPS have significantly smoother motor actuation, and simply by virtue of having been regularized for smooth actions, achieve comparable error tracking performance to the vanilla PPO agents, with less than 20\% peak motor usage.
        Agents trained with CAPS, on just tracking error minimization, are flight-worthy, posing limited risk for a real-world test, something that would otherwise likely not be achievable with vanilla PPO.
        As noted by Koch et al.~\cite{NFThesis} however, the simulated environment does not fully capture the dynamics of the drone.
        Achieving stable real-world flight required a baseline thrust (of roughly 30\%) to compensate for gravity.
        We account for this by re-introducing a minimum thrust reward.
        Training with CAPS completes successfully within 1 million time-steps, constituting a 90\% reduction in data intensity and $8\times$ wall-time speedup over Neuroflight.
        
        Results shown in Table~\ref{Drone-flight-table} compare the performance on mean absolute error (MAE), average current draw and smoothness of CAPS-optimized agents against a tuned PID controller (tuned with the Ziegler-Nichols method\cite{Ziegler1942OptimumSF}), the best Neuroflight agent, and PPO agents trained with exclusively temporal or spatial smoothing regularization.
        Fig.~\ref{Drone-FFT} similarly shows a comparison of the FFTs of the control signals over the same agents.
        Variance statistics are computed over 10 independently trained agents, trained with different random seeds and initialization.
        When compared to PID and Neuroflight, we incur a loss in tracking accuracy, though this was within acceptable limits for real-world flight, with a mean overall error of under 10 deg/s. 
        {CAPS-optimized} agents however are notably smoother, with virtually no high-frequency control components.
        Consequently, they consume significantly less power, lower even than the PID controller.
        Importantly, the tracking performance of CAPS agents is more consistent between simulation and reality, which we believe is also partially due to the simpler reward structure it enables. 
        This allows us to achieve 100\% repeatability with our training pipeline on this task. 
        
        Deploying the training pipeline also enabled an ablation study on the real-world impact of temporal and spatial smoothing. 
        Spatial smoothing was expected to enable smoother control but to come at the cost of forcing the output signals into distinct bands, but with better sim-to-real tracking parity. 
        Temporal smoothing was expected to offer more smoothness than unregularized training and likely better tracking but was expected to be more sensitive to domain shift. 
        Our results support these hypotheses and also show that both regularizers used together in CAPS offer superior performance in training and in transfer. 
        As we were not able to gain access to code from Shen et al.~\cite{shen2020deep} and knowing that RL performance can be very implementation dependent~\cite{Engstrom2020Implementation}, we cannot provide a thorough comparison of our method against theirs. 
        Based on the theory however, we expect their method to work similarly to our spatial smoothing term, and as we demonstrate, spatial smoothing alone is not enough for practical policy transfer.
    
        \begin{table}[t]
          \caption{Flight performance on Drone}
          \label{Drone-flight-table}
          \centering
          \setlength\tabcolsep{1.9pt}
          \begin{tabular}{c||c|c|c}
            \hline
            {Agent} & {MAE (deg/s)} $\downarrow$ & Current (Amps) $\downarrow$ & Sm $\cdot 10^3$ $\downarrow$ \\ \hline \hline
            \multicolumn{4}{c}{Simulated Validation} \\ \hline
            {PID}          & $11.41$   & -NA-  & $0.29$    \\ 
            {Neuroflight}  & $7.30$   & -NA-   & $3.23$      \\ \hline
            {PPO + Temporal}  & $11.36 \pm 2.19$   & -NA-   & $0.056 \pm 0.007$      \\ 
            {PPO + Spatial}  & $16.66 \pm 4.57$   & -NA-   & $0.021 \pm 0.007$      \\ 
            {PPO + CAPS}  & $9.26 \pm 1.03$  & -NA-  & $0.021 \pm 0.012$  \\ \hline \hline
            \multicolumn{4}{c}{On-Platform Live Test-Flights} \\\hline
            {PID}          & $5.01$      &  $8.07$  & $0.4$ \\
            {Neuroflight}  & $5.19$      &  $22.87$ &  $4.3$  \\ \hline
            {PPO + Temporal}  & $7.82 \pm 2.42$   & $ 7.59 \pm 2.24 $   & $ 1.10 \pm 0.32 $       \\ 
            {PPO + Spatial}  & $14.85 \pm {6.85}$   & $4.59 \pm 2.70 $ & $ 0.37 \pm 0.22 $      \\ 
            {PPO + CAPS}   & $9.28 \pm 2.31$  & $4.86 \pm 2.32 $ & $0.16 \pm 0.02$  \\ \hline
            \end{tabular}
            \vspace{-\baselineskip}
        \end{table}
        
        We believe that the main utility of CAPS is that it allows the problem of policy smoothness to be addressed entirely at the level of the RL learning algorithm, requiring no intervention with the training environment.
        The intuitively derived formulation also makes it easier to understand what behavior the terms affect and encourage, making it easier to tune that when using reward-engineering.
        CAPS also exposes an interesting and unintuitive element of neural-network based control.
        Having sampling/actuation rates be several times that of the cutoff frequency in a system's dynamical frequency response is typical and helpful as it helps prevent aliasing when sampling and allows for the disambiguation of noise from dynamics.
        We find however, that allowing neural networks to actuate at rates significantly faster than the system dynamics appears to make them more prone to learn non-smooth behavior.
        These are also however the cases with more potential for improving control smoothness, as in the case of the pendulum, lander and also our quadrotor drone.
    
    \paragraph*{\textbf{Why not just use a filter?}}\label{nofilter}
    
    
        It is natural to question why one would not just filter the RL policies' outputs.
        In principle, the use of filters may seem like a good idea --- filters have known benefits in classical controls systems~\cite{DSP}.
        Unfortunately, neural network-based control behaves fundamentally differently and we found no existing literature that studies the effects of filtering on RL-based control.
        Neural network policies are not typically trained with integrated filters, so deploying them with a filter changes the dynamical response expected by the network and could result in anomalous behavior, as we demonstrate in Fig.~\ref{FilterToy}. 
        Preliminary results suggest that na\"ively introducing filters into the training pipeline, without also introducing the corresponding state history can result in a catastrophic failure to learn, \rev{likely because this} breaks the Markov assumption which forms the basis of RL theory.
        On the other hand, \rev{compensating by} including the state history \rev{to ensure the Markov assumption is satisfied} could significantly increase the representation complexity of the problem due to the higher dimensional input states
        \rev{and CAPS shows that this compromise is not always necessary.}
        While we had some success in training RL with filtered outputs and state histories on simple tasks, we were not able to successfully do so on the drone, likely due to increased complexity in system dynamics.
        Classical techniques offer many potential tools to improve the behavior of controllers, but more thorough investigation on the benefits and drawbacks of these tools is needed before applying them to learned control. 

\section{Conclusion}
    Understanding the behavioral characteristics of a controller is a crucial element of its utility, yet we find that work studying the fundamental behavior of RL-based controllers is limited.
    By carefully considering how RL policies can be optimized with action smoothness in mind, we developed Conditioning for Action Policy Smoothness (CAPS).
    CAPS trains for two aspects of action smoothness: (i) sequential actions should be smooth, and (ii) similar states should map to similar actions.
    CAPS can be implemented independently of any reward engineering and changes to the training environment, allowing it to be added to practically any continuous control RL algorithm.
    Our results demonstrate the general utility of CAPS in improving the smoothness of learned policies without significant negative impact on the policies' performances.
    \rev{This learned smoothness is also} transferable between domains --- as shown when transferring attitude control agents trained in simulation to the real quadrotor.
    Practically, this resulted in an almost 80\% reduction in power consumption and significantly reduced heating and wear of motors during flights while maintaining good tracking of input controls.
    We hope future work can use our method and its implications on the value of regularization in RL as a springboard for developing more robust controllers.
    \vspace{.8\baselineskip}
    
    \noindent \textbf{Acknowledgements:} This work was supported in part by grants NSF 1724237 and NSF CCF 2008799.
    We also thank Weifan Chen for helping to run experiments.

\bibliographystyle{IEEEtran}
\bibliography{references.bib}

\end{document}